Karolina Rudnicka

University of Freiburg

# Variation of sentence length across time and genre: influence on the syntactic usage in English

**Abstract**

The goal of this paper is threefold: i) to present some practical aspects of using full-text version of Corpus of Historical American English (COHA), the largest diachronic multi-genre corpus of the English language, in the investigation of a linguistic trend of change; ii) to test a widely held assumption that sentence length in written English has been steadily decreasing over the past few centuries; iii) to point to a possible link between the changes in sentence length and changes in the English syntactic usage. The empirical proof of concept for iii) is provided by the decline in the frequency of the non-finite purpose subordinator *in order to*. Sentence length, genre and the likelihood of occurrence of *in order to* are shown to be interrelated.

**Keywords:** sentence length; language change; semicolon; mega-corpora; Late Modern English, *in order to*

1. Introduction

Sentence length, defined as the number of words that come between the opening word (starting with a capital letter) and the end punctuation sign such as full stop, question mark or exclamation mark is a subject that attracts a lot of interest in different contexts, such as cognitive linguistics, rhetoric and language teaching, to name just a few. There are handbooks of English suggesting language users and language learners use shorter sentences, as this would make their English appear more modern (McGregor 2002: 33), or advising them to keep their sentences within a given length range (Hult 2015: 105):

> In an age of tweets, text messages and *Facebook* posts, English sentences seem to be decreasing in length. Sentences in English vary from one word (*Help!*) to forty or more words. To keep your sentences both readable and meaningful, however, you will want to stay in the fifteen- to twenty-word range most of the time.

Another frequently investigated context is the relation between sentence length and syntactic complexity (e.g. Klare 1963: 170; Westin 2002: 81; Biber & Conrad 2009: 152; Fahnestock 2011: 169; Štajner & Mitkov 2012: 1578), which might directly and indirectly influence text complexity and text comprehension (Fries 2010: 21; Fahnestock 2011: 169-170). Sentence length is one of the main variables which are included in various readability formulas (equations yielding quantitative scores assessing text difficulty) such as the Gunning Fog Index and Lexile reading measure (Gross et al. 2002: 171; Fahnestock 2011: 170). One of the applications these measures have is, for instance, in the assessment of school textbooks for certain age groups. Similarly, the length of sentences produced by the language users is used as a valuable index of language development in children (e.g. Davis 1937: 69).

From a diachronic perspective the relation between sentence length and syntactic complexity can tell us something about changes in the way in which authors use the resources offered by the linguistic system (Biber & Conrad 2009: 152). Or it could help us answer the question asked by Štajner and Mitkov (2012: 1577-1578) whether texts are becoming simpler and easier to read, as sentence length is frequently used as one of several factors of text complexity (the other ones are e.g. Automated Readability Index, sentence complexity and the use of passive voice).

The present paper looks at sentence length from a different angle and focuses on its evolution across the last two hundred years, offering insights into possible explanations for the observed decrease. Additionally, it points to a possible influence this decrease might have on the syntactic usage of English. The first section presents works focused on the evolution of sentence length across time and, in the same context, addresses the question of the influence of changing punctuation conventions. The second section presents the methodology of working with full-text version of COHA. It contains a detailed description of the possible pitfalls of using textual versions of mega-corpora and introduces *Mathematica* as a programme of choice for the study.

In total, the lengths of more than nineteen million sentences extracted from the corpus are calculated. The results of the comprehensive analysis reveal a visible decrease in sentence length across time for all of the investigated genres. It is suggested that this decrease can have an influence on the syntactic usage of constructions that show correlating patterns of frequency development (decrease or increase in the frequency of use). The third part is devoted to the non-finite purpose subordinator *in order to* which serves as an example construction and which is shown to be decreasing in the frequency of use. This decrease, however, is different in the case of each of the investigated genres. Insights in the history of *in order to* and the genre-related distribution of *in order to* are then presented to justify the hypothesis that the general decrease in sentence length and the general decrease in frequency of use of *in order to* might be interrelated. It is likely that the described phenomenon is generalizable to more constructions from the network of English purpose subordinators such as *in order that*, *so as to*, *lest* (Rudnicka 2018, in preparation) or to other constructional networks of both English and other languages.

## 2. Sentence length in written English – the diachronic evolution across genres

More than one hundred years ago, Lewis (1894: 34) stated that *"the English sentence has decreased in average length at least one-half in three hundred years"*. His work presents "a count of the average number of words to the sentence and to the paragraph, in representative authors since the middle of the fifteenth century". Although the main focus of Lewis's paper is on the structure of a paragraph in written English across centuries, he additionally identifies a visible trend towards a decrease in average sentence length. According to his calculations, the paragraph in the late 19th century has the same length as in the 16th century, but it contains twice as many sentences (1894: 170). On the other hand Säily, Vartiainen and Siirtola (in press) who look at sentence length evolution across the time period represented by the *Parsed Corpus of Early English Correspondence* (c.1410–1681) find that the sentence length stayed roughly the same in the entire period studied. This might mean that the decrease in sentence length referred to by Lewis either starts a bit later or is only visible in literary works and not in private correspondence.

Biber and Conrad (2009: 151-153) compare linguistic characteristics of novels from the 18th and 20th century and call the change in syntactic complexity "perhaps the most important change" in the way that authors use the resources of the given linguistic system. Sentence length is one of the measures used to study the differences in syntactic complexity between the works by particular authors from the eighteenth and twentieth centuries. The comparison of sentence lengths in narratives across the two historical periods yields results pointing towards a rather steady decrease of the average sentence length (Biber & Conrad 2009: 152):

> While there is some variation at any given historical period, there is also a very steady progression from the extremely long sentences of Defoe to the short sentences of Vonnegut and Bellow.

The change in sentence length is said to be, to a large extent, a reflection of changing punctuation practices, such as "a much more extensive use of colons and semicolons in earlier historical periods" (Biber & Conrad 2009: 153).

The work of Gross, Harmon and Reidy (2002) focuses on scientific prose. Their results show "a definite shrinking in average sentence length over time, from 33 words in 1876-1900 to 30 words in 1901-1925 to 27 in 1976-2000" (2002: 171), while the clausal density is said to have remained relatively stable. Additionally, their work offers an important observation suggesting than the decrease in sentence length might actually be a phenomenon generalizable to other languages. The comparison of average sentence lengths across languages in 19th century passages shows there is a decrease in English, German and French scientific texts (2002: 124). Gross et al. (2002:171) identify two opposite trends which concern the readability of the scientific prose of today. According to their conclusions, on the one hand it is becoming more difficult to read because of the use of increasingly complex and compact noun phrases.

On the other hand, it is becoming easier to read "because of its declining sentence length and number of clauses per sentence". The work of Dorgeloh (2005) on patterns of agentivity and narrativity in early science discourse, adds a few more observations on the development of language of science from the 15th century to the present, and on the factors that shaped the way scientific texts look like today. In her work (2005: 92) she claims that "modern science texts tend to nominalise the experience and to impersonalise the argument", whereas in the past it was typical for the narrative and the argument to be based on personal reference. The observations concerning the more complex and compact noun phrases and the decrease in the number of clauses per sentence are further supported by the research of Hundt et al. (2012: 224), which also focuses on scientific discourse. Additionally, their research notes that the marked decrease in the frequency of relative clauses coincides with a marked decrease of sentence length occurring around the beginning of the twentieth century (2012: 225).

      A large part of the recent research dealing with sentence length decrease focuses on newspaper language. Westin (2002) provides statistically significant evidence for this effect in a wide range of English newspapers, such as *The Times*, *Guardian* and *Daily Telegraph* for the period 1900-2000. Additionally to the sentence length itself, she also investigates sentence length distributions in the texts from particular years. One of the observations she makes is that "The shortest sentences, from 1 to 10 words, increased in number, from 5.7% in 1900 to 16.6% in 1993" (Westin 2002: 81). Westin concludes that if we assume there is a relation between sentence length and sentence complexity, "it is obvious that the complexity of the sentences in English upmarket newspaper editorials decreased considerably during the 20th century" (Westin 2002: 81). Likewise, focusing on newspaper language, Fries (2010) shows a decrease of approximately 10 words in sentence length during the 18th century in the *London Gazette*. Apart from the diachronic development of sentence length, he compares sentence length in different sections of the investigated journal and points to sections that, on average, seem to show longer sentences such as foreign news, and sections that tend to contain more short sentences, such as the advertising section. Schneider (2002: 98) reports that the sentence length in newspaper news has decreased by an average of 15 words since 1700. The reasons for this decrease given by Schneider (2002: 98ff.) include a necessity for greater comprehensibility for a mass readership and a trend in which newspaper language became more similar to the spoken word. Also Westin (2002: 161) concludes that authors of some of the studied newspapers might on purpose adjust the language of their editorials in order to attract a broader audience than competing newspapers.

*2.1 Just a matter of punctuation conventions?*

> English punctuation will apparently never stop causing division among scholars. There appears to be serious disagreement about its nature, functions and formal status.
> (Shou 2007: 195)

Whereas most authors of the empirical studies described in the section above claim that the decrease in sentence length in written English is, at least partly, a linguistic fact, other scholars, e.g. Fahnestock, express a view that the observed decrease might actually be "a by-product of changing punctuation conventions" (2011: 265) or a reflection of changing punctuation practices (Biber & Conrad 2009: 153). The full stop, exclamation mark and question mark are said to have taken over the work that was previously also done by colons, semicolons and commas. As Westin (2002: 79) points out, some historical linguists "consider the colon and semicolon as sentence delimiters". Support for this observation is provided by e.g. Miller who in his paper offers a first-hand view on punctuation practices at the beginning of the twentieth century (1908: 327):

> There are four structural equivalents of the period; namely, the semicolon (or colon), the structural connective, the series, and the balance.

This might suggest that the assumed decrease in sentence length would, to say the least, look differently if one used semicolons and colons as sentence delimiters, instead of only using full stops, exclamation marks and question marks. Westin (2002: 79) claims that in her earlier research she adopted both perspectives and the two approaches have shown to have almost identical patterns of development (1997: 16-17; quoted in Westin 2002: 79). Thus, in the paper in which she shows the universal decrease in sentence length in *The Times*, *Guardian* and *Daily Telegraph* for the period 1900-2000, she includes only sentences ending with a full stop, a question mark, or an exclamation mark. The question arises if the two approaches compared by Westin in her first work would still show similar patterns of development if she did her research also on some slightly earlier texts. Since super-long sentences, containing 60 or more words, seem to be much more typical for the nineteenth century than for the twentieth century (Biber

& Conrad 2009: 152-153), the inclusion of earlier texts in the investigation could have changed the observed patterns of development.

The results from COHA seem to support this observation, as semicolons were much more frequently used in the beginning of the nineteenth century, than they were in the twentieth century. Figure 2-1 shows the development of the frequency (normalised, per million words) of semicolons across time. The data used for the plot were obtained via the online interface of COHA.[1] We see a dramatic decrease in the frequency of use of semicolons happening between 1810 and 1900. From 1900 on, there still is a decrease, but the curve is much less steep, suggesting that the beginning of the twentieth century was a turning point for the use of semicolons.

Still, even though colons and semicolons were once considered sentence delimiters, the punctuation of present-day English follows different rules. If we accept the view that at least part of the decrease in sentence length can be attributed to a change in punctuation practices, does it mean that this decrease is only apparent? And, if the previous punctuation system was comprehensive why did it not stay the same until today?

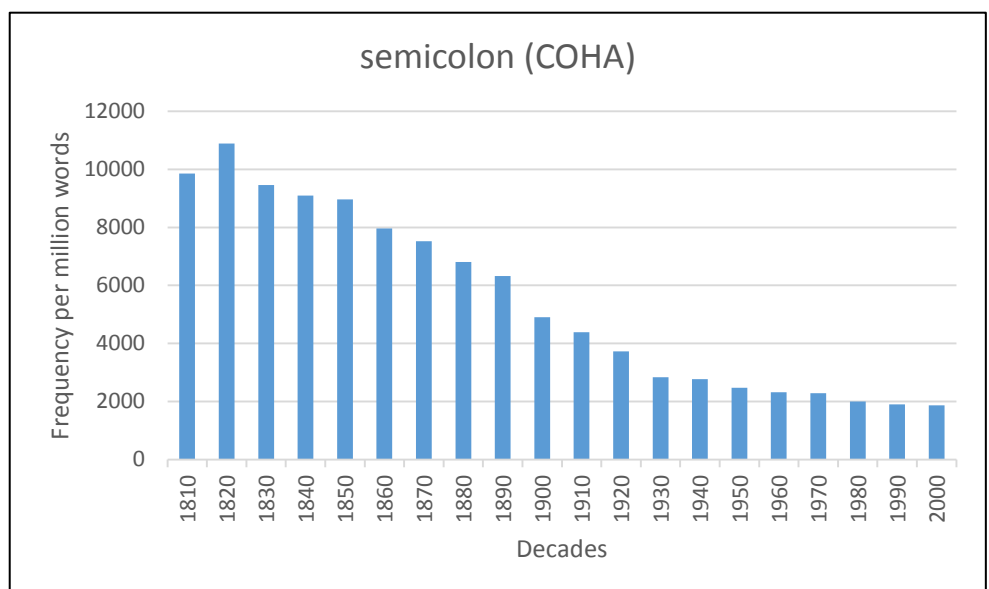

Fig. 2-1: Frequency of semicolons across the period 1810 – 2009 in COHA[2].

The explanations for the decrease in sentence length offered by Schneider (2002: 98ff) and Westin (2002:161) seem to make a lot of sense, also in the context of changing punctuation practices. Adding to their conclusions one could hypothesise that as the development of mass readership went on, the newspaper editors and authors might have adopted different punctuation conventions because of the need for greater text comprehensibility for the new society of readers. The dramatic decrease in the frequency of use of semicolons (Fig. 2-1) coincides with the achievement of mass literacy by the American society and the invention of new printing technologies, since these developments took place around the mid- to late nineteenth century, as observe e.g. Hames and Rae (1996: 227):

> All this changed with technological innovations in the mid- to late nineteenth century, and it is at this time that the rise of the independent mass circulation daily newspaper began. In addition to changes in printing technology that made production easier, the invention of telegraph and the advent of the wire services meant that news could be quickly transmitted to all areas of the country.

---

[1] http://corpus.byu.edu/coha/old/
[2] Data retrieved from COHA on May 13th 2017.

One of the possible explanations for the change in punctuation conventions might thus have been the development of mass readership. The change in punctuation conventions might have, in turn, led to a decrease of sentence length in written present-day English.

3. A comprehensive analysis of sentence length in the time period of 1800-2000

This work defines sentence length as the number of words that come between the opening word starting with a capital letter and the end punctuation sign, namely a full stop, exclamation mark and question mark. The main aim of the analysis is to detect and visualise the trend of change in the sentence length across the time period 1810-2009 and across different genres of American English. The novelty of this analysis lies in the application of full-text COHA data. Since most of the research described in the section above is based on relatively small samples of texts, it is hoped that the use of a large collection of textual data available as COHA corpus sheds more light on the phenomenon and makes an interesting contribution to the discussion about sentence length in the English language. Also, the practical aspects and technical details of using full-text COHA are expected to be of interest to scholars who would like to profit from the availability of big data in their research.

*3.1 Design of the analysis and methodology*

*3.1.1 Full-text COHA*

COHA is known as the largest diachronic multi-genre corpus of American English and it contains 400 million words of text from the time period 1810-2009 belonging to four genres, namely *fiction*, *magazine, newspaper* and *non-fiction*. For the purposes of this analysis the contents of *fiction* genre are modified in order to decrease the level of bias, namely, a subgenre of *movie & play script* is removed from *fiction* and treated as an independent genre – a proxy for the spoken language. According to the terminology in Culpeper and Kytö (2010) the data constituting *movie & play script* section are speech-purposed, i.e. designed in order to imitate real-time spoken interaction.

COHA is available online via an interface at https://corpus.byu.edu/coha/. It is, however, impossible to conduct an in-depth analysis of the evolution of sentence length via the online interface, as the length of search strings in COHA cannot exceed the limit of fifteen words. Because of this fact, the paper uses the full-text offline version of the corpus. The full-text COHA has a form of twenty folders. Each of the folders encompasses one decade and contains various amounts of text files (txt format) representing different genres. The name of each file serves the purpose of genre and decade identification, e.g. mag_2007_387216.txt belongs to the *magazine* genre and is from the 2000 decade, while the last number in the file name may be used to identify its exact source in the sources_coha.xls table. All in all, the full-text COHA contains 116 615 text files.

*3.1.2 Genres in COHA*

On the webpage of COHA it can be read that the corpus is well-balanced in terms of genres, and the proportions of genres stay roughly the same from decade to decade. What are the contents of particular genres? For the purpose of clarity, a few words need to be said about the COHA genres themselves.

The genre accounting for the largest share of COHA (48-55% of the total in each decade) is *fiction*. According to the information on the webpage, *fiction* contains texts from digitized books from many different sources, such as *Project Gutenberg* (1810 – 1930) and *The Cornell University Library Making of America Collection* (1800 – 1900), a digital library of primary sources in American social history[3]; scanned books (1930 – 1990); movie and play scripts which can also be found in COCA[4] another mega-corpus of the same family.

*Magazine* contains digitized journals from *The Cornell University Library Making of America Collection* (1810 – 1900), scanned magazines and the contents of the COCA genre *popular magazine* (1990 – 2009), which is described as "nearly 100 different magazines, with a good mix (overall, and by year) between specific domains (news, health, home and gardening, women, financial, religion, sports, etc.)".[5] Some examples of the contemporary magazines include *Time*, *Men's Health*, *Good Housekeeping*, *Cosmopolitan*, *Fortune*, *Christian Century*, *Sports Illustrated*. According to a detailed list of COHA sources, which is attached to the full-text version of the corpus, among the magazines representing the early nineteenth century are: *The North American Review*, *The New England Magazine*, *The United States Democratic Review*, *New Englander* and *Yale Review*.

*Newspaper* genre contains various digitized texts from *The New York Times* (decades 1860 – 2009), *The Chicago Tribune* (1900 – 2009), *The Wall Street Journal* (1910 – 1989), *The Christian Science Monitor* (1930 – 2009), *The Boston Globe* (1980 – 2009), *USA Today* (1990 – 2009), *The San Francisco Chronicle* (1990 – 2009), *The Washington Post* (1990 – 2009), *The Atlanta Journal Constitution* (1990 – 2009), *The Houston Chronicle* (1990 – 2009), *The Associated Press* (1990 – 2009), *The Denver Post* (1990 – 2009).

As one can read on the webpage of COHA, in the *non-fiction* genre there are ebooks from *Project Gutenberg* and www.archive.org (for the decades 1810 – 1900), scanned books (1900-1990) and the contents of COCA (1990 – 2009). The texts in the *non-fiction* genre are balanced across the Library of Congress classification system[6], which in practice means that in the detailed list of all the text sources, there is a letter assigned for each source, which classifies it as belonging to one of the twenty one categories such as *A – general works*; *B – philosophy, psychology, religion*; *C – auxiliary sciences of history*; etc. The genre *non-fiction* is, however, not homogenous across the whole time span of COHA. Until the decade 1990 it is mostly composed of books, and starting from the decade 1990 texts from nearly a hundred different peer-reviewed journals such as *Ear, Nose & Throat Journal*; *The Physical Educator*; *Arab Studies Quarterly* are added to the corpus. On the COHA webpage these texts are referred to as the contents of COCA (in COCA they constitute *Academic Journals* genre). As one can see in the COHA online interface, the genre *non-fiction* is sometimes referred to as *non-fiction* and sometimes as *non-fiction books*. In the present work it is treated as the rough equivalent of the *learned* category from the Brown Corpus[7].

---

[3] http://ebooks.library.cornell.edu/m/moa/
[4] http://corpus.byu.edu/coca/
[5] http://corpus.byu.edu/coca/old/
[6] http://www.loc.gov/catdir/cpso/lcco/

*3.1.3 Sentence tokenisation – methodology*

The present study uses *Mathematica* as the programme of choice to tokenise texts into individual sentences and provide the word counts. The version 10.4, which is used here, contains tools designed especially to deal with linguistic data.[8] Below the description of the sentence tokenisation procedure with some notes on the content recognition features of the programme:

1. The corpus data (text files) are imported and stored in the form of strings of text. Each text file becomes one string, so that it is possible to control the origin of the text (decade and genre). The tool TextSentences[9] parses strings of texts into individual sentences creating a very long list of sentences. The precision of TextSentences tool can be assessed as 93-98%, which means that it correctly tokenises between 93% and 98% of all the sentences. The exact precision depends to a large extent on the quality of a given text sample, which is in general worse for the earlier decades (1810 – 1890) and better for the later decades (1900 – 2000).

2. As explained on the COHA webpage, the texts included in the full-text version have been modified in order to avoid any economic impact on the holders of the copyright. In practice this means that out of every 200 words of the running text, ten words have been removed and replaced with "@". The modifications have thus a form of "@ @ @ @ @ @ @ @ @ @". The distribution of modifications is the same for each decade and genre, it is also independent of the text source. Keeping these strings in the dataset would make the analysis of sentence lengths meaningless, as we do not know, for instance, if there is a full stop somewhere in these modified part of text. All the sentences containing the string "@ @ @ @ @ @ @ @ @ @" are thus removed from our list of sentences. As explained in the COHA tutorial, due to the random distribution of the modifications, the removal of affected data should not have an influence on the statistical analysis.

3. The TextSentences tool recognizes common abbreviations such as "Mr.", "Mrs." or "Jr.", so we do not need to worry about a situation in which the full stops here would be treated as sentence delimiters.

---

[7] The Standard Corpus of Present-Day Edited American English
[8] http://reference.wolfram.com/language/guide/LinguisticData.html
[9] https://reference.wolfram.com/language/ref/TextSentences.html

4. On the other hand, cases of enumerations, used frequently to mark the beginning of a chapter / article / passage and containing a full stop exemplified by (2) are not detected by the programme and only a careful visual inspection of the data could keep them out of the dataset. Given the size of the corpus the best solution is to remove all the one-word sentences from the final dataset.

(2) Art.1 (1839, NF, American Fruit Garden, COHA)
Vol. II. p. 77. (1870, MAG, Atlantic, COHA)

5. Also the common tags such as "<P>" need to be removed from the data.
6. Parsing is conducted for each text file of the corpus leaving us with over nineteen million sentences (19,768,290 to be precise). Table 3.1 shows the exact numbers of sentences extracted for each decade and genre.

| Decade | Magazine | Newspaper | Non-fiction | Fiction | Movie & play script |
|---|---|---|---|---|---|
| 1810 | 2,242 | NA | 11,299 | 44,843 | NA |
| 1820 | 43,080 | NA | 42,524 | 150,977 | NA |
| 1830 | 78,267 | NA | 88,951 | 299,450 | NA |
| 1840 | 92,112 | NA | 109,403 | 368,834 | NA |
| 1850 | 116,018 | NA | 93,772 | 385,595 | NA |
| 1860 | 121,195 | 9,048 | 92,201 | 442,831 | NA |
| 1870 | 125,968 | 34,773 | 96,158 | 513,287 | NA |
| 1880 | 130,737 | 48,199 | 109,512 | 544,721 | NA |
| 1890 | 137,595 | 52,954 | 107,512 | 549,291 | NA |
| 1900 | 164,386 | 63,440 | 116,223 | 655,739 | 49 |
| 1910 | 206,582 | 62,465 | 126,706 | 705,553 | 5,720 |
| 1920 | 258,497 | 154,553 | 117,524 | 741,621 | 33,877 |
| 1930 | 254,573 | 138,671 | 113,920 | 709,298 | 70,225 |
| 1940 | 262,520 | 134,679 | 115,432 | 736,662 | 74,429 |
| 1950 | 266,879 | 145,524 | 117,991 | 740,315 | 68,921 |
| 1960 | 260,855 | 148,992 | 118,577 | 722,116 | 83,936 |
| 1970 | 253,159 | 141,422 | 113,998 | 756,625 | 68,288 |
| 1980 | 267,081 | 168,057 | 122,086 | 800,548 | 78,927 |
| 1990 | 335,913 | 191,881 | 115,085 | 871,569 | 82,290 |
| 2000 | 360,496 | 182,950 | 121,762 | 995,354 | NA |
| **In total** | 3,738,155 | 1,677,608 | 2,050,636 | 11,735,229 | 566,662 |

Table 3-1: Number of sentences extracted per decade and per genre.

*3.2 Results*

A visual inspections of a sample of a data set from both the earlier and later decades led to an assumption that potential bias in any direction caused by the errors of the algorithm (a string of sentences not parsed, undetected short bogus sentences, the presence of undetected abbreviations etc.) is more or less the same for each decade. Thus, it is still possible to reliably study the diachronic evolution of the sentence length, even though one shall be careful when providing "exact" values of variables such as the aforementioned mean sentence length.

WordCount tool is used to provide the word counts of each sentence in each genre. After running the routine we obtain a very long list of pairs (a, b), where "a" refers to the decade labelling of a particular sentence and "b" to its word length. The 19 768 290 data points (one for each sentence) are used to create a bigger picture of the evolution of sentence length for each of the genres across time. Figures 3-1 to 3-5 present visualisations of the trends in the form of box plots. The white line on each of the boxes indicates the median, the black line, which is also shorter than the white one, indicates the mean.

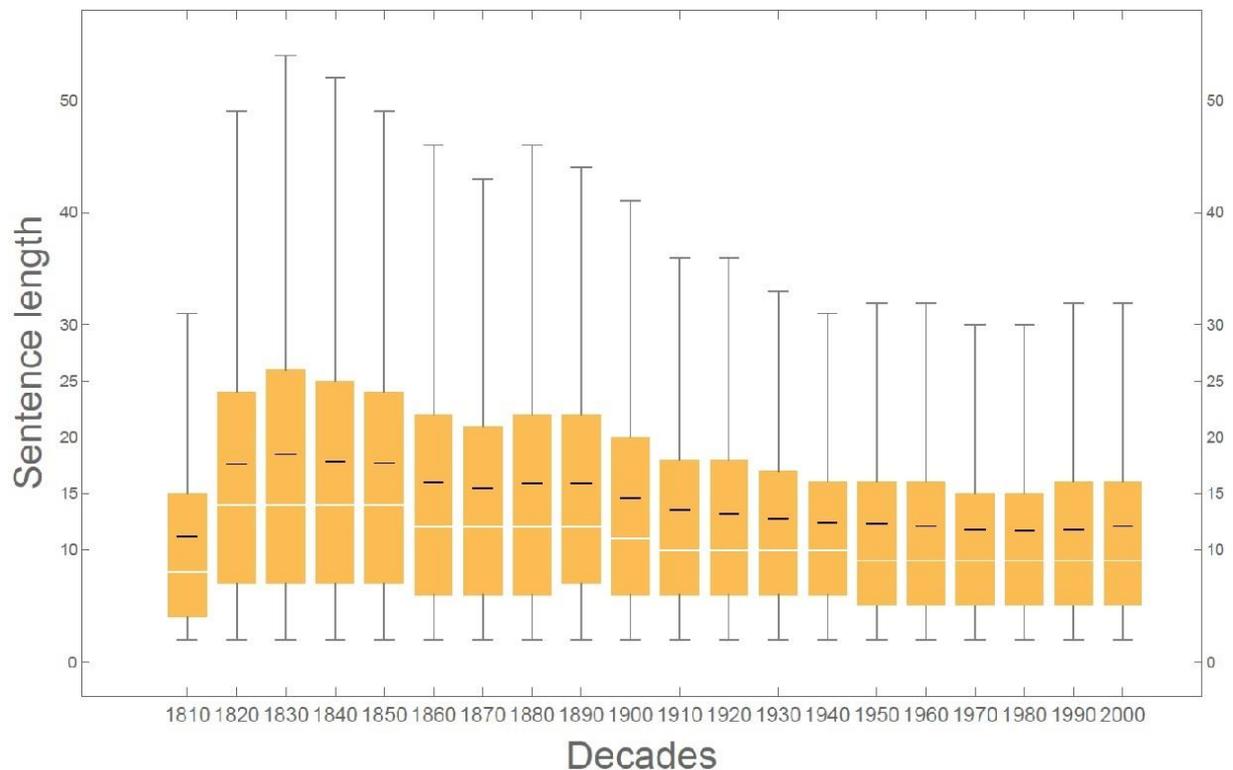

Figure 3-1: Sentence length across the time period 1810-2009 for the COHA genre *fiction*.

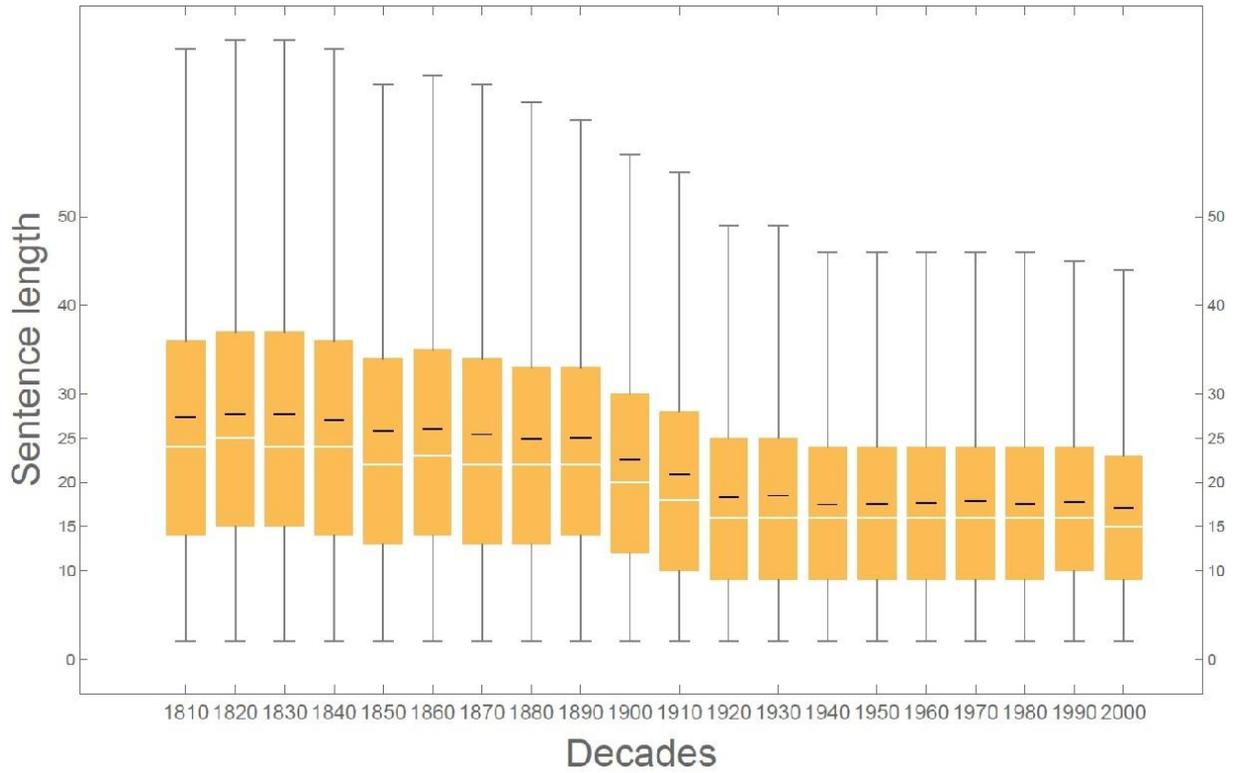

Figure 3-2: Sentence length across the time period 1810-2009 for the COHA genre *magazine*.

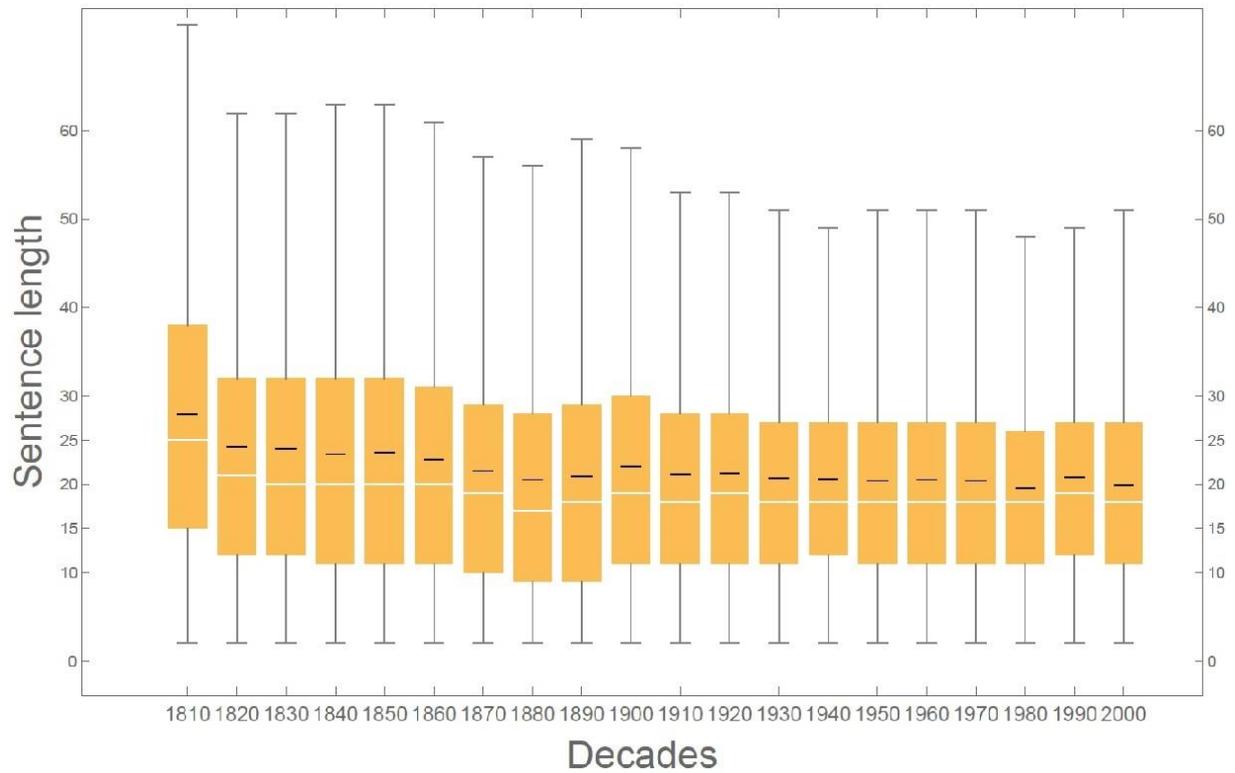

Figure 3-3: Sentence length across the time period 1810-2009 for the COHA genre *non-fiction*.

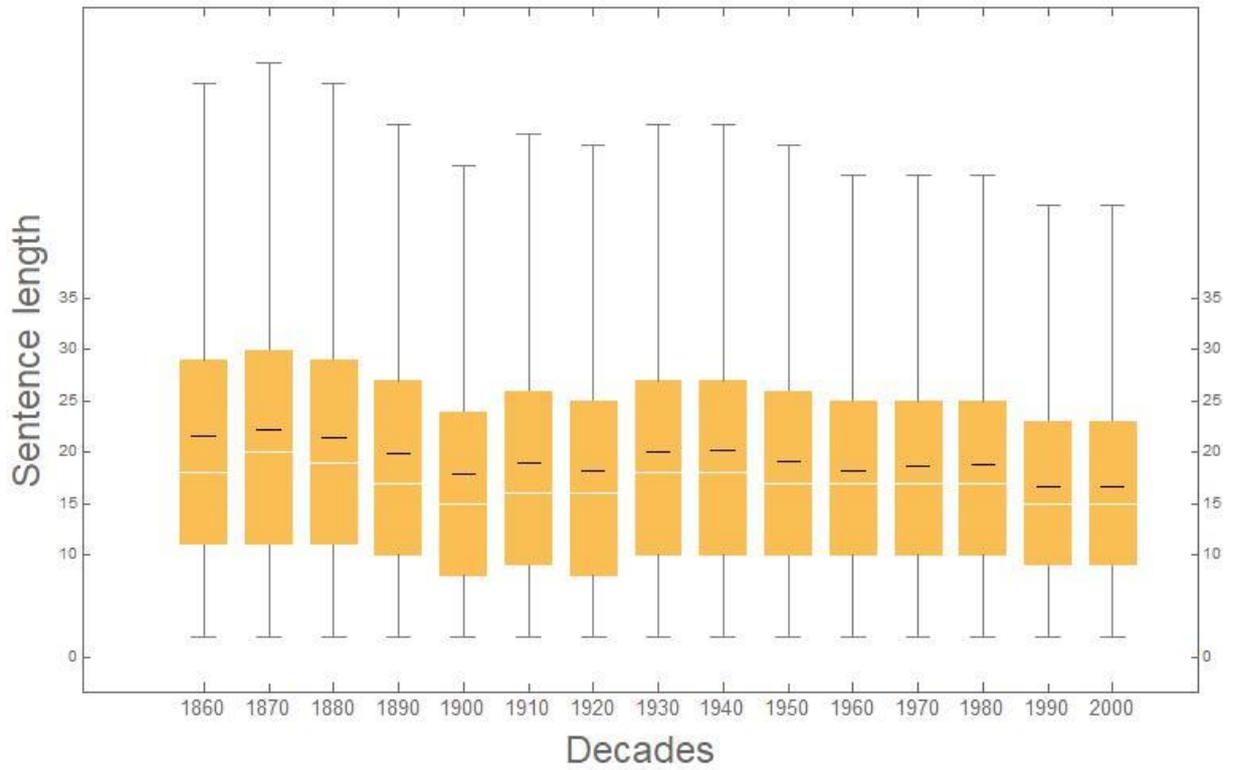

Figure 3-4: Sentence length across the time period 1860-2009 for the COHA genre *newspaper*.

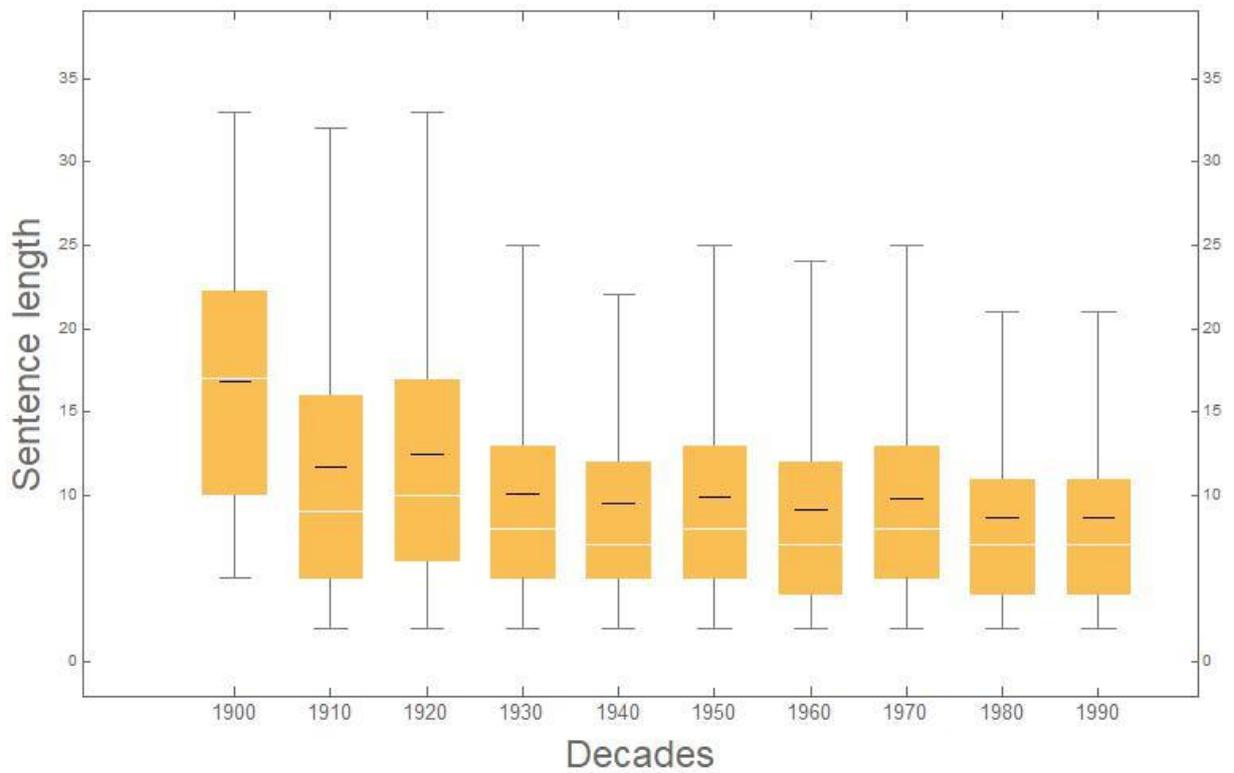

Figure 3-5: Sentence length across the time period 1900-1999 for the COHA subgenre *movie & play script*.

The plots in Figures 3-1 to 3-5 reveal a decrease in the sentence length for each of the genres. Figure 3-1 presents the distribution of sentence lengths for *fiction*. We can see a visible downward trend starting from 1830. Also in the case of *magazine* (Fig. 3-2) there is a visible decrease of sentence lengths across time. Sentence length in *non-fiction* (Fig. 3-3) also decreases, however, the degree of decrease does not seem to be that visible as in the case of *fiction* and *magazine*. In the case of *newspaper* (Fig. 3-4) there is as well a decrease, however, it is worth noting that the starting decade is 1860. COHA does not feature texts belonging to this genre from earlier decades. Similarly, for *movie & play script* (Fig. 3-5) the starting decade is 1900. Even though we have less data representing this subgenre of *fiction*, we can assume that there is also a certain degree of sentence length decrease over time.

The value which will be used to compare the results between the studied genres is the mean value of sentence length. Table 3-2 gives an overview of the mean values. As we can see, if we use the mean value to draw conclusions, there is a visible decrease over time for each of the five studied genres. For *magazine*, the sentences in the onset of the twenty-first century are approximately 10 words shorter than in the beginning of the nineteen century, for *non-fiction* the difference amounts to 8 words. Texts in *newspaper* are almost 5 words shorter now than they were in the middle of the nineteenth century. In the case of *fiction*, there first seem to be an increase in the sentence length and then a visible decrease starting from the third decade (1830). The increase in sentence length in 1810 and 1820 might seem surprising, but a closer inspection of the dataset for 1810 and 1820 reveals a large amount of two-word sentences such as in (3):

(3) Eugenia Nothing . (COHA: 1817; FIC; "How to try a lover")
    Prince Mentzkioff . (COHA: 1812; FIC; "Alexis, the Czarewitz")

The number of two-word sentences in *fiction* per million words is almost three times higher in 1810 than in 2000 decade. Neither *magazine*, nor *non-fiction* of 1810 come close to the amount found in *fiction*. It seems that the bias can be attributed to characteristics of texts included in the dataset for 1810 and, possibly, to some degree of 'pollution' of the texts.

Also the proxy for the spoken language – *movie & play script* – noted a decrease in the mean sentence length. It is, though, harder to asses even the approximate level of decrease, as for this genre we have the smallest dataset. Not surprisingly, it is also the genre with the shortest sentences nowadays – in the nineties its mean sentence length was 8.6 words, whereas for *magazine* and *newspaper* it was 17.78 and 16.27 respectively. Long sentences seem to be the domain of *non-fiction*. The mean sentence length in this genre amounts to almost 20 words and, as we can see in Fig. 3-3, the upper whiskers of the box plots representing recent decades reach values higher than in the case of any other genre.

| **Decade** | *Magazine* | *Newspaper* | *Non-fiction* | *Fiction* | *Movie&play script* |
|---|---|---|---|---|---|
| **1810** | 27,29 | NA | 27,96 | 11,12 | NA |
| **1820** | 27,77 | NA | 24,23 | 17,64 | NA |
| **1830** | 27,76 | NA | 24,06 | 18,51 | NA |
| **1840** | 27 | NA | 23,41 | 17,85 | NA |
| **1850** | 25,84 | NA | 23,58 | 17,69 | NA |
| **1860** | 26,02 | 21,57 | 22,77 | 15,93 | NA |
| **1870** | 25,44 | 22,13 | 21,52 | 15,47 | NA |
| **1880** | 24,96 | 21,39 | 20,5 | 15,9 | NA |
| **1890** | 25,02 | 19,9 | 20,9 | 15,9 | NA |
| **1900** | 22,65 | 17,83 | 21,97 | 14,58 | 16,86 |
| **1910** | 20,9 | 18,89 | 21,1 | 13,54 | 11,66 |
| **1920** | 18,34 | 18,14 | 21,25 | 13,2 | 12,46 |
| **1930** | 18,53 | 19,96 | 20,74 | 12,69 | 10,1 |
| **1940** | 17,54 | 20,2 | 20,56 | 12,37 | 9,49 |
| **1950** | 17,57 | 19,04 | 20,36 | 12,36 | 9,84 |
| **1960** | 17,72 | 18,2 | 20,53 | 12,09 | 9,12 |
| **1970** | 17,87 | 18,72 | 20,45 | 11,76 | 9,77 |
| **1980** | 17,65 | 18,77 | 19,65 | 11,73 | 8,63 |
| **1990** | 17,78 | 16,72 | 20,8 | 11,8 | 8,62 |
| **2000** | 17,14 | 16,7 | 19,94 | 12,07 | NA |

Table 3-2: Mean sentence lengths in words in COHA across time and genre.

*3.3 Discussion*

The observed decrease in sentence length across time seems to be an outcome of many factors. Some, already mentioned in this work, range from the need for greater text comprehensibility accompanying the development of mass readership (Schneider 2002: 98ff), to competition for the potential reader (Westin 2002: 161) and change of punctuation conventions. Contrary to the idea that the decrease of sentence length is only a "by-product" of change in punctuation practices (Fahnestock 2011: 265), it did not stop once the punctuation became its present-day version. So there have to be other trends and phenomena at work. One of them might be the overall informalization of the language of the media – a trend described by Leech et al. (2009: 239). The decrease in sentence length is, however, not only typical for the media language but, as this work aims to show, also for other genres such as *fiction* and *non-fiction*. Thus, it might actually be one of the symptoms of a larger phenomenon concerning all the genres of the written language. A phenomenon encompassing this kind of change is the trend introduced by Mair as "colloquialization" (1998: 153), namely "a significant stylistic shift in twentieth century English" (Mair 2006: 187) due to which the written language becomes more similar to the spoken language and more tolerant to various degrees of informality. One of the clear examples of colloquialization is the increase in the frequency of use of contractions (negative contractions in particular) during the twentieth century, which has been really dramatic in written American English (Mair 2006: 190). The results of the present analysis, however, also point towards a decrease in the sentence length in *movie & play script* genre, which is used as a proxy for the spoken language. Genre conventions typical for scripts might account for the fact that *movie & play script* has the shortest sentences overall. They, however, do not explain the gradual decrease in sentence length. If the decrease of sentence length in written language might be a symptom of colloquialization what is it a symptom of when happening in spoken language? The answer to this question might be of interest to not only linguists but also to sociologists and cognitive scientists. However, to be answered in a satisfactory way, it requires a separate study. For the *non-fiction* genre, part of the decrease might have to do with the trend associated with the growing lexical difficulty of the scientific prose, which make the word-load of the sentences considerably lower (Gross et al. 2002; Dorgeloh 2005; Hundt et al. 2012).

## 4. Sentence length and syntactic usage

According to the results of the previous analysis, sentence length in American English has decreased quite considerably during the last 200 years. But does this mean that the messages produced by the language users of today are less complex? Probably not. On the other hand, a decrease in sentence length will, quite likely, have an influence on the contents of this new, shorter sentence.

An interesting case in this context is provided by a non-finite subordinator *in order to*. The first use of its present-day version dates back to the early seventeenth century (according to OED Online).[10] *In order to* came into existence as a kind of 'reinforcement' of the purposive meaning of the *to*-infinitive (Schmidtke-Bode 2009: 174). According to Los (2005: 28), "the function in which the *to*-infinitive first appeared was that of a purpose adjunct". As the scope of usage of the *to*-infinitive extended greatly during the Old and Middle English periods, the addition of *in order* in front of *to* was supposed to disambiguate the purposive meaning.

*In order to* is, even today, one of the main purposive subordinators. Figure 4-1 shows its frequency of use (normalised, per million words) in COHA. The frequency values look rather stable until the end of the nineteenth century. But then, just around the turn of century, we can see a visible decrease. This decrease follows a pattern very similar to the decrease of sentence length presented in Figures 3-1 to 3-5.

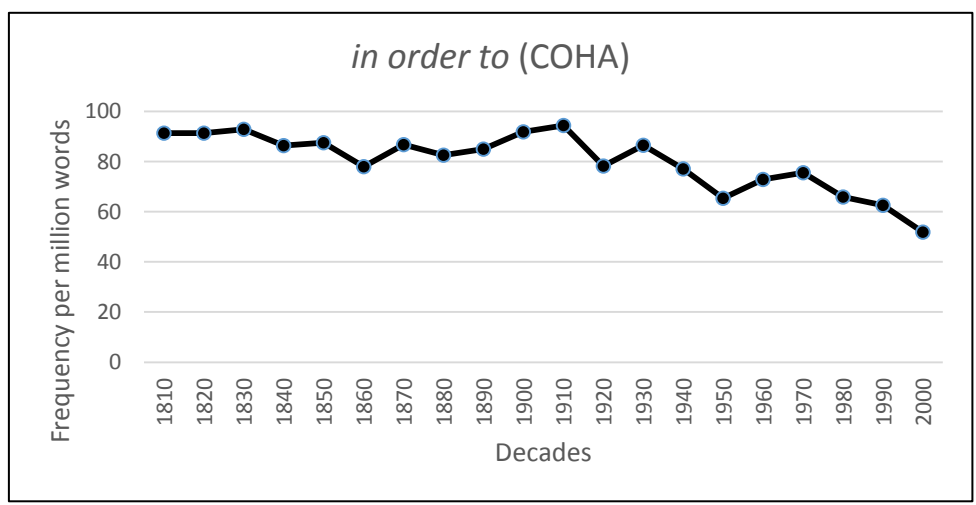

Figure 4-1: Frequency of *in order to* across the time period 1810 to 2009[11].

---

[10] OED Online, s.v. *in order to*, retrieved on May 29, 2016 from http://www.oed.com
[11] Data retrieved from COHA on May 30th 2016.

This correlation between the decrease of frequency of *in order to* and the decrease of sentence length could reflect a process opposite to what happened when *in order* was added in front of the purposive *to*-infinitive. Since sentences are shorter now, chances are that there are less sentences containing more than one *to*-infinitive, so the need for precision, to the fulfilment of which *in order to* was born, might just not be there anymore.

Another observation that supports the link between the decrease in sentence length and the decrease in the frequency of use of *in order to* is the fact that this decrease seems, at least to some extent, genre-dependent. Figure 4-2 presents the decrease in frequency of *in order to* across time and across different genres of COHA (normalised frequency, per million words).

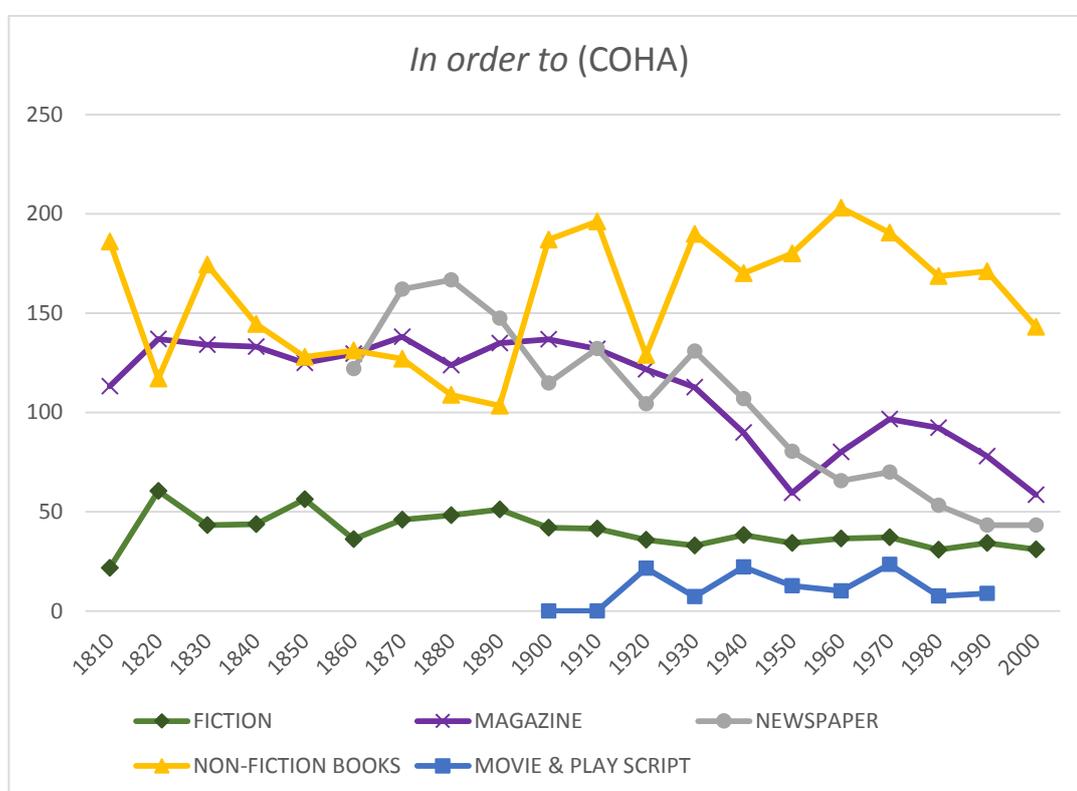

Figure 4-2: Frequency per million words of *in order to* across the time period 1810 to 2009[12] in *fiction*, *magazine*, *non-fiction*, *newspaper* and *movie & play script* (time period 1860 to 2009).

As we can see *non-fiction* is the only genre in which we can see some degree of increase in the frequency of *in order to*. Part of the explanation for this might be provided by the fact that, at present, it is the genre with the longest sentences among all the investigated genres (see Table 3-2). In *fiction*, *newspaper* and *magazine* there is a visible decrease in the frequency of *in order to*. The genre *movie & play script* is the genre with the lowest overall number of instances.

---

[12] Data retrieved from COHA on e.g 15th 2017.

Another part of the explanation might be in the original function of *in order to*. Since its aim was to provide more clarity and precision, the genre *non-fiction* might actually prove to become its "natural environment" for now due to the rhetorical conventions and the need for clarity and unambiguousness typical for *learned* genres.

## 5. Conclusions

The results of the comprehensive analysis of sentence length across time (1810 – 2000) and genre show a decrease in sentence length for all the investigated genres. The degree of the decrease is slightly different for each genre, namely, in *fiction* sentences are approximately (on average) 6.5 words shorter now than they were in the beginning of the nineteenth century, for *magazine* the decrease amounts to 10.2 words, for *newspaper* 4.9, for *non-fiction* 8 and for *movie & play script* 8.3 words. The initial sentence lengths were, however, different for each genre, with *movie & play script* starting from the lowest sentence length and finishing at the lowest sentence length from all genres and *non-fiction* starting and staying the genre with longest sentences.

Various explanations have been offered to account for the observed decrease in sentence length across time and genre. Among the processes at work which might influence sentence length there are informalization of the language of the media (Leech et al. 2009: 239), colloquialization (Mair 1998: 153), a trend towards less explicitness in writing (noted by e.g. Biber & Gray 2010), trends accounting for changes in the scientific discourse such as the increase in the use of compressed and complex noun phrases (Gross et al. 2002: 171; Dorgeloh 2005: 92; Hund et al. 2012: 236) and a decrease in the relative clause frequency (Gross et al. 2002: 171; Hundt et al. 2012: 225). In the *non-fiction* genre, the decrease happening between 1990 – 2009 might, at least partially, be attributed also to the addition of COCA contents (mostly texts from peer-reviewed academic journals).

The more marked decrease in sentence length observed around the beginning of the twentieth century might have to do with the development of mass readership (Schneider 2002, quoted in Fries 2010: 23) and competition for the potential reader (Westin 2002: 161) and a change in punctuation conventions (noted by e.g. Fahnestock 2011: 256 and Biber & Conrad 2009: 153) which as such might actually be the resultant of the two previous phenomena. The above mentioned processes and developments might account for the decrease in sentence length across particular written genres but they do not explain the observed decrease in the proxy for the spoken language – *movie & play script*. The question whether the observed decrease would also be visible in the spoken language and which processes could possibly account for it remains open.

The paper aimed to point out that the decrease in sentence length might be linked to changes in the syntactic usage of English. Non-finite purpose subordinator *in order to* is used to illustrate the potential influence that the decrease of sentence length might have on the constructional layer of the language. The observed decrease in sentence length is, according to the present work, an instantiation of a higher-order phenomenon. What are other examples of higher-order phenomena? According to Hilpert (2013: 14) processes and phenomena which affect multiple constructions at the same time represent higher-order phenomena, among the examples there are the development of a global, phonotactic constraint and deflexion (a general loss of inflectional morphological categories):

> [W]henever several members of grammatical paradigm are in demise or even whole groups of paradigms disappear, there are reasons to view the change as non-constructional, because a higher level of grammatical organization than the construction is concerned.

There are thus reasons to view the decrease of sentence length as a higher-order phenomenon at work, which, in turn, might, at least to some extent, be influencing the syntactic usage. The shorter sentences of today might make the use of more explicit purpose subordinators such as *in order to* redundant. It is very likely that *in order to* is not the only construction affected, as the decrease in the frequency of use is generalizable to other constructions of the network of purpose subordinators, such as *so as to*, *lest*, *in order that* (Rudnicka 2018, in preparation). A scenario for the future of *in order to* could be a situation in which it becomes a marker of style, used mostly in *learned* texts or for rhetorical purposes.

Since the decrease in sentence length has been, at least in the language of science (Gross et al. 2002: 124), shown to be generalizable to other languages, it seems worthwhile to check if other genres across different languages and across time would show comparable degrees of sentence length decrease, and if the possible explanations for this decrease would be similar cross-linguistically.

**Language Corpora**

Davies, Mark. (2008-). *The Corpus of Contemporary American English (COCA): 520 million words, 1990-present*. Available online at http://corpus.byu.edu/coca/

Davies, Mark. (2010-). *The Corpus of Historical American English (COHA): 400 million words, 1810- 2009*. Available online at https://corpus.byu.edu/coha/

**References**

Biber, Douglas & Conrad, Susan. 2009. *Register, Genre, and Style*. Cambridge: Cambridge University Press.

Biber, Douglas & Gray, Bethany. 2010. Challenging stereotypes about academic writing: Complexity, elaboration, explicitness. *Journal of English for Academic Purposes* 9, 2-20.

Davis, Edith A. 1937. Mean sentence length compared with long and short sentences as a reliable measure of language development. *Child Development* 8(1): 69-79.

Dorgeloh, Heidrun. 2005. Patterns of agentivity and narrativity in early science discourse. In *Opening windows in discourse and texts from the past,* Janne Skaffari, Matti Peikola, Ruth Carroll, Risto Hiltunen & Brita Warvik (eds), 83-94. Amsterdam & Philadelphia: Benjamins.

Fahnestock, Jeanne. 2011. *Rhetorical Style: The Uses Of Language In Persuasion*. New York: Oxford University Press.

Fries, Udo. 2010. Sentence Length, Sentence Complexity and the Noun Phrase in 18[th]-Century News Publications. In *Language Change and Variation from Old English to Late Modern*

Variation of sentence length across time and genre